\documentclass{article}

\PassOptionsToPackage{numbers, compress}{natbib}

\usepackage[preprint]{neurips_2026}

\usepackage[utf8]{inputenc} 
\usepackage[T1]{fontenc}    
\usepackage{hyperref}       
\usepackage{url}            
\usepackage{booktabs}       
\usepackage{amsfonts}       
\usepackage{nicefrac}       
\usepackage{microtype}      
\usepackage{xcolor}         
\usepackage{amsmath}
\usepackage{amssymb}
\usepackage{booktabs}
\usepackage{graphicx}
\usepackage{eso-pic}        
\title{LMM-Track4D: Eliciting 4D Dynamic Reasoning in LMMs via Trajectory-Grounded Dialogue}

\author{%
  Chaoyue Li$^{1,5}$\thanks{Equal contribution.} \quad
  Yongxue Xu$^{2,5\ast}$ \quad
  Jie Feng$^{3,5\ast}$ \quad
  Jiayu Ding$^{4,5}$\thanks{Corresponding author.} \\
  \\
  $^1$Huazhong University of Science and Technology \quad
  $^2$Sun Yat-sen University \\
  $^3$Beihang University \quad
  $^4$Peking University \quad
  $^5$InkMind.AI \\
  \texttt{hustlichaoyue@hust.edu.cn} \quad \texttt{jyding25@stu.pku.edu.cn}
}

\begin{document}

\AddToShipoutPictureBG*{
  \AtPageUpperLeft{
    \put(50,-50){\includegraphics[height=0.65cm]{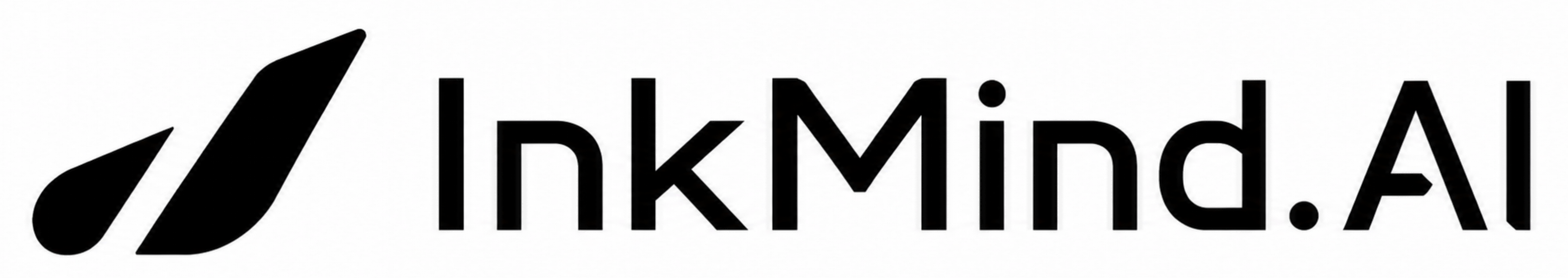}}
  }
}

\maketitle

\begin{abstract}
Recent large multimodal models (LMMs) have become increasingly capable on image and video understanding, yet still struggle to sustain 4D continuous spatiotemporal dynamic reasoning. To study this capability gap, we formulate trajectory-grounded multi-turn spatiotemporal dialogue, a new task in which a model must answer spatiotemporal queries while returning structured 3D target trajectories over an entire short clip or a specified segment of a longer clip, and introduce Track4D-Bench, a benchmark with 526 clip-level dialogue samples spanning 23.5k frames and 7.5k object annotations, for training and evaluation. Building on this task, we propose LMM-Track4D, which combines RTGE (Ray--Time Geometry Encoding), a dedicated streaming state token TRK for long-horizon dynamic propagation, and an Object-Slot Kinematic, Residual-Anchor (OSK-RA) decoder for stable 4-step 3D state estimation under occlusion and viewpoint variation. Experiments on Track4D-Bench show consistent improvements over strong baselines, suggesting that explicit dynamic state modeling is a useful design principle for eliciting 4D dynamic reasoning in LMMs. Our code and dataset will be publicly available at https://github.com/mikubaka88/LMM-Track4D.
\end{abstract}

\section{Introduction}
\label{sec:intro}
Recent advances have greatly improved the image- and video-understanding ability of large multimodal models (LMMs)~\cite{jin2026tirflowactivevideosearch,jin2026vistamitigatingsemanticinertia,jin2025videomemenhancingultralongvideo,jin2026himachierarchicalmacromicrolearning,jin2026dgpodistributionguidedpolicy,wen2025ai,wang2025accelerating,wen2026innovator,ke2026flash,wen2026evostreaming}, and recent systems have begun to extend these gains toward 4D-aware reasoning~\cite{wangCompositional4dDynamic2025,zhouVLM4DSpatiotemporalAwareness2025,zhouUni4DLLMUnifiedSpatioTemporalaware2025,zhou2025llava,huangThinkingDynamicsHow2026a}. Yet they remain weak at sustained \textbf{4D dynamic reasoning} across time, viewpoints, and multi-turn interaction: the challenge is not recognizing isolated observations, but maintaining a physically coherent account of evolving scene state as new evidence arrives.

\begin{figure*}[t]
\centering
\includegraphics[width=\textwidth]{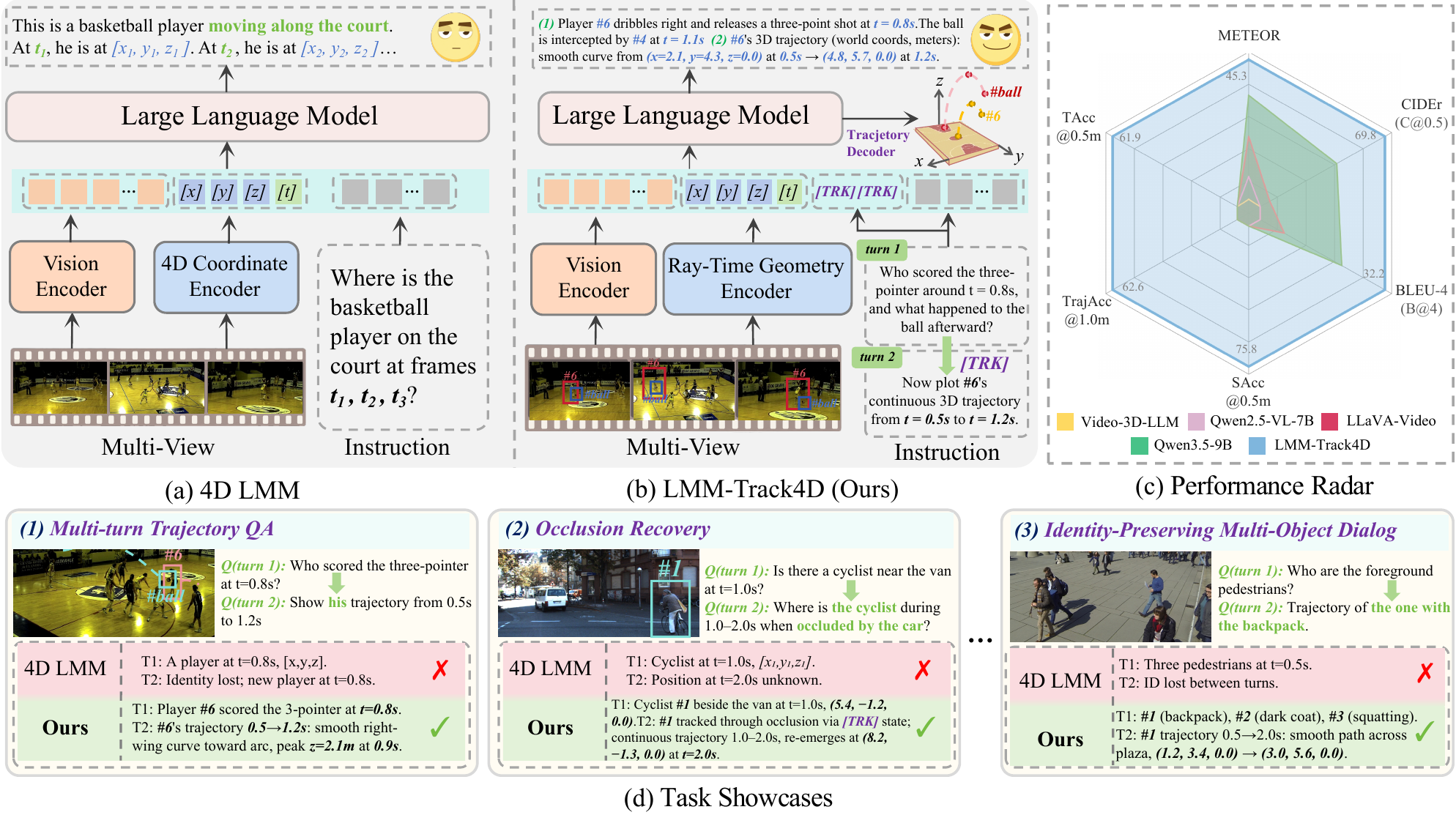}
\caption{Task and model overview for trajectory-grounded multi-turn spatiotemporal dialogue. A generic 4D LMM lacks explicit geometry-grounded state maintenance, whereas LMM-Track4D adds RTGE, persistent TRK propagation, and structured trajectory decoding. Representative cases and the performance radar highlight the gap on dialogue-grounded 4D dynamic reasoning.}
\label{fig:task_overview}
\end{figure*}

To study this gap, we formulate \textbf{trajectory-grounded multi-turn spatiotemporal dialogue}, where a model answers spatiotemporal questions while, when required, returning structured 3D trajectories over a short clip or a local segment of a longer interaction. This task couples language reasoning with explicit state prediction, complementing recent work on dynamic video reasoning and 4D-aware multimodal modeling~\cite{huangThinkingDynamicsHow2026a,yanVISAReasoningVideo2025,yinMLLM4DVisualbasedSpatialtemporal2026a}, and directly tests whether a model can maintain and revise scene dynamics throughout dialogue. Figure~\ref{fig:task_overview} contrasts a generic 4D LMM with our stateful design and illustrates representative dialogue-grounded 4D dynamic reasoning cases.

This setting remains difficult because most current LMM pipelines are optimized to interpret already observed content~\cite{liBLIP2BootstrappingLanguageimagea,liVideoChatChatcentricVideo,zhang2024llavanext-video,wangInternVL35AdvancingOpensource2025}, not to preserve explicit dynamic state. Video is still represented largely as framewise tokens, geometry remains weakly modeled, and dialogue history lacks a dedicated mechanism for identity or motion continuity; as a result, a model may appear competitive on 4D-aware perception while still failing to maintain coherent scene dynamics over time.

To make this problem trainable and measurable at scale, we introduce \textbf{Track4D-Bench}. Built through a structured data generation pipeline rather than direct raw-video prompting, Track4D-Bench assembles 526 auditable clip-level dialogue samples from APIDIS, KITTI, and WildTrack~\cite{castanedo2011multi,geiger2013vision,chavdarova2018wildtrack}, covering 23.5k frames and 7.5k object annotations, alongside verified 3D trajectories, object-centric visual evidence, synchronized keyframes, textual descriptions, and multi-turn supervision. Its short-clip regime combines full-clip trajectory outputs with qualitative spatiotemporal questions, whereas its long-clip regime pairs local-segment tracking with single-moment queries. This design enables systematic evaluation of whether an LMM can sustain 4D dynamic reasoning across complementary temporal scopes and answer formats.

Building on this task and benchmark, we propose \textbf{LMM-Track4D}, designed to maintain and update dynamic state throughout interaction. It combines RTGE (Ray--Time Geometry Encoding) to inject geometry and time into visual tokens, a persistent TRK state token to preserve continuity across turns, and an Object-Slot Kinematic, Residual-Anchor (OSK-RA) decoder to recover stable 3D trajectories from multiview evidence. Together, these components target the main bottlenecks of continuous 4D dynamic reasoning: geometry grounding, cross-turn continuity, and robust state decoding.

Across \textbf{Track4D-Bench}, LMM-Track4D improves trajectory-grounded 3D state estimation and long-horizon dialogue consistency, with especially strong gains in dense non-keyframe inference, severe occlusion, and long-range identity preservation. These results indicate that eliciting 4D dynamic reasoning in LMMs is not simply a matter of scaling video perception or generic video question answering (QA); it benefits from task formulations, supervision, and architectures that explicitly model evolving scene dynamics over time. Our contributions are summarized as follows:
\begin{itemize}
\item We identify a key limitation of current LMMs in sustaining 4D dynamic reasoning across time, viewpoints, and multi-turn interaction, and formulate \textbf{trajectory-grounded multi-turn spatiotemporal dialogue} to study it.
\item We introduce \textbf{Track4D-Bench}, a benchmark built through a structured 3D trajectory--visual evidence--dialogue pipeline, with short-clip and long-clip regimes for training and evaluating 4D dynamic reasoning.
\item We present \textbf{LMM-Track4D}, which integrates RTGE, a streaming state token TRK, and an Object-Slot Kinematic, Residual-Anchor (OSK-RA) decoder to preserve and update dynamic information across views, time, and dialogue turns.
\item Experiments on Track4D-Bench show gains in dense non-keyframe inference, long-horizon identity preservation, and occlusion robustness, highlighting the value of explicit dynamic state modeling for eliciting 4D dynamic reasoning in LMMs.
\end{itemize}

\section{Related Work}
\label{sec:related}

\textbf{Video large multimodal models.} Video LMMs have advanced from early vision-language systems such as BLIP-2~\cite{liBLIP2BootstrappingLanguageimagea} and VideoChat~\cite{liVideoChatChatcentricVideo} to stronger video assistants including LLaVA-NeXT-Video~\cite{zhang2024llavanext-video}, Video-LLaVA~\cite{linVideoLLaVALearningUnited2024}, and InternVL3.5~\cite{wangInternVL35AdvancingOpensource2025}. General multimodal models such as GPT-4o~\cite{hurst2024gpt}, Gemini 2.5~\cite{comanici2025gemini}, and Qwen3-VL~\cite{bai2025qwen3} further improve open-domain video reasoning. Recent 4D-oriented systems, including LLaVA-ST~\cite{liLLaVASTMultimodalLarge2025a}, LLaVA-4D~\cite{zhou2025llava}, VLM4D~\cite{wangCompositional4dDynamic2025}, Uni4D-LLM~\cite{zhouUni4DLLMUnifiedSpatioTemporalaware2025}, Thinking in Dynamics~\cite{huangThinkingDynamicsHow2026a}, and MLLM-4D~\cite{yinMLLM4DVisualbasedSpatialtemporal2026a}, move closer to spatiotemporal reasoning. A remaining limitation is persistent multi-turn 4D reasoning, in which entity state must be maintained and revised across interaction.

\textbf{3D-aware multimodal grounding and reasoning.} SpatialVLM~\cite{chenSpatialVLMEndowingVisionlanguage2024}, OmniSpatial~\cite{jiaOmniSpatialComprehensiveSpatial2025}, Flatland~\cite{zhangFlatlandSpaceTeaching2025a}, and Spatial-MLLM~\cite{wuSpatialMLLMBoostingMLLM2025} inject explicit spatial structure into vision-language models. Learning Videos in 3D~\cite{zhengLearningVideos3D2025a} and LLaVA-3D~\cite{zhuLLaVA3DSimpleEffectivea} improve cross-view and 3D grounding, while Chat-Scene~\cite{huang2024chat}, 3D-LLaVA~\cite{deng3DLLaVAGeneralist3D}, and 3D-R1~\cite{huang3DR1EnhancingReasoning2025} extend multimodal reasoning in 3D environments. A remaining limitation in this setting is dynamic state grounding, where geometry must support dialogue-conditioned identity, motion, and state updates over time rather than only static or short-horizon 3D understanding~\cite{ding2026extrinsplatdecouplinggeometrysemantics,ding20263dinstructionambiguitydetection}.

\textbf{Dynamic scene benchmarks.} Related settings include MeViS~\cite{dingMeViSLargescaleBenchmark2023}, VISA~\cite{yanVISAReasoningVideo2025}, PointOdyssey~\cite{zhengPointOdysseyLargescaleSynthetic2023}, and Dynamic Stereo~\cite{karaevDynamicStereoConsistentDynamic2023}. Other works study richer 4D supervision through Stereo4D~\cite{jinStereo4DLearningHow}, Motion4D~\cite{zhouMotion4DLearning3Dconsistent}, and Understanding Dynamic Scenes in 4D~\cite{huangUnderstandingDynamicScenes2025}. Track4D-Bench instead couples verified 3D supervision with multi-turn dialogue on APIDIS~\cite{castanedo2011multi}, KITTI~\cite{geiger2013vision}, and WildTrack~\cite{chavdarova2018wildtrack}. A remaining benchmark gap is query-conditioned, stateful 4D reasoning, in which language understanding and auditable 3D state maintenance must remain consistent across turns.

\section{Track4D-Bench}
\label{sec:benchmark}

\subsection{Benchmark Construction}

\begin{figure*}[t]
\centering
\includegraphics[width=\textwidth]{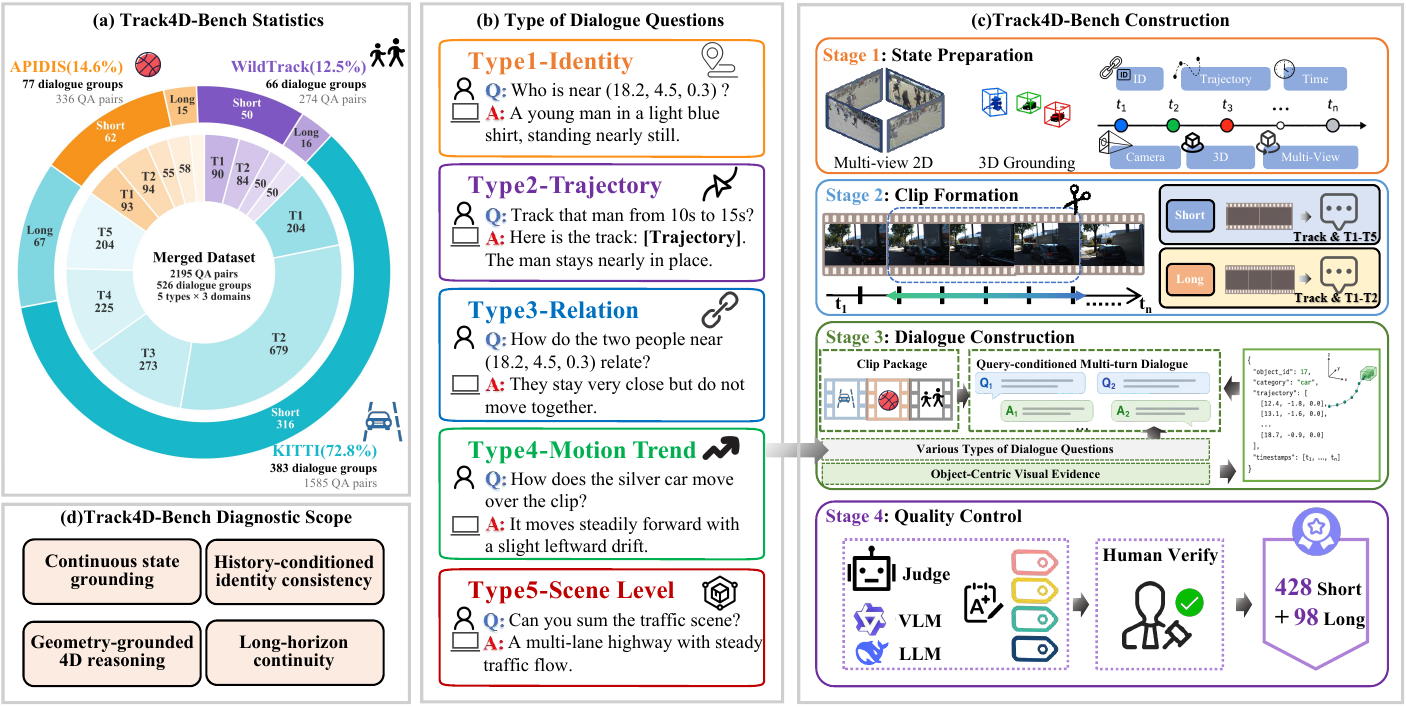}
\caption{Track4D-Bench overview. The figure reports benchmark scale, question taxonomy, the 3D trajectory--visual evidence--dialogue construction pipeline, and the main diagnostic axes targeted by evaluation. Together these components make 4D dynamic reasoning trainable, measurable, and diagnostically interpretable.}
\label{fig:track4d_benchmark}
\end{figure*}

Track4D-Bench is introduced to make trajectory-grounded multi-turn spatiotemporal dialogue trainable, measurable, and diagnostically interpretable at scale. Each instance links explicit dynamic state, a clip-level temporal unit, query-conditioned dialogue, and quality control, so that language and geometric supervision remain auditable and failures can be attributed to state maintenance, geometry grounding, and history-conditioned reasoning rather than uncontrolled data noise. The benchmark is built through four stages---state preparation, clip formation, dialogue construction, and quality control---whose detailed implementations are deferred to Appendix~\ref{sec:appendix_benchmark_construction}. The current release contains 526 clip-level dialogue samples from APIDIS, KITTI, and WildTrack~\cite{castanedo2011multi,geiger2013vision,chavdarova2018wildtrack}, covering 23.5k frames and 7.5k object annotations. Short clips pair full-clip trajectory outputs with qualitative spatiotemporal questions, whereas long clips pair broader temporal context with local-segment tracking and single-moment queries. Figure~\ref{fig:track4d_benchmark} summarizes the benchmark scale, question taxonomy, auditable construction pipeline, and diagnostic axes used to probe 4D dynamic reasoning.

\subsection{Evaluation Protocol}

Track4D-Bench is evaluated at the dialogue-turn level because the task mixes qualitative language responses with explicit 3D outputs. Non-trajectory turns are scored by CIDEr, BLEU-4, and METEOR~\cite{vedantam2015cider,papineni2002bleu,banerjee2005meteor}. For geometric turns, we compute point error $e_i = \|\hat{\mathbf{p}}_i - \mathbf{p}_i\|_2$ and trajectory-step error $e_{j,t} = \|\hat{\mathbf{p}}_{j,t} - \mathbf{p}_{j,t}\|_2$, with mean trajectory error $\bar{e}_j = \frac{1}{|\mathcal{V}_j|}\sum_{t \in \mathcal{V}_j} e_{j,t}$.

The main geometric metrics are
\[
\mathrm{SAcc}@0.5 = \frac{1}{|\mathcal{T}_{\mathrm{pt}}|} \sum_{i \in \mathcal{T}_{\mathrm{pt}}} \mathbf{1}[e_i < 0.5], \qquad
\mathrm{Traj\mbox{-}Acc}@1.0 = \frac{1}{|\mathcal{T}_{\mathrm{traj}}|} \sum_{j \in \mathcal{T}_{\mathrm{traj}}} \mathbf{1}[\bar{e}_j < 1.0],
\]
\[
\mathrm{TAcc}@0.5 = \frac{\sum_{j \in \mathcal{T}_{\mathrm{traj}}} \sum_{t \in \mathcal{V}_j} \mathbf{1}[e_{j,t} < 0.5]}{\sum_{j \in \mathcal{T}_{\mathrm{traj}}} |\mathcal{V}_j|}.
\]
All geometric errors are computed in world-coordinate $(x,y,z)$ space. Details on valid-step handling, multi-target alignment, and corpus-level micro averaging are deferred to Appendix~\ref{sec:appendix_impl_details}.

\subsection{Benchmark Significance and Diagnostic Scope}

Track4D-Bench is a query-driven and explicitly supervised testbed for continuous 4D dynamic reasoning. Because each turn is tied to structured state and visual evidence, it evaluates whether a model can maintain \textbf{physically coherent state} rather than produce semantically plausible but weakly grounded answers. Qualitative turns are measured by CIDEr, BLEU-4, and METEOR, while explicit geometric turns are measured by SAcc@0.5, Traj-Acc@1.0, and TAcc@0.5, so language quality and 3D state accuracy are assessed under a \textbf{unified dialogue-turn protocol}.

More importantly, the benchmark is structured around \textbf{identity consistency}, \textbf{geometry-grounded motion inference}, and \textbf{long-horizon continuity}. Multi-turn queries reveal whether target state remains stable under occlusion, distractors, and viewpoint change, while synchronized keyframes and explicit 3D supervision expose cross-view grounding, non-keyframe inference, and partial-visibility reasoning as measurable point or trajectory errors. The short regime emphasizes full-clip trajectory grounding with qualitative dialogue, whereas the long regime stresses local-state reasoning in broader temporal context, together testing whether a model can preserve coherent dynamic state over extended interaction and motivating the state-maintenance design in Section~\ref{sec:method}.

\section{Method}
\label{sec:method}

\subsection{Overview}

Given a clip observed from $V$ views at physical timestamps $\mathcal{T} = \{\tau_1, \ldots, \tau_M\}$, let $I_{v,\tau}$ denote the red-green-blue (RGB) image from view $v \in \{1, \ldots, V\}$ at time $\tau \in \mathcal{T}$, and let $K_{v,\tau}$ and $E_{v,\tau}$ denote the corresponding camera intrinsics and extrinsics. At dialogue turn $k$, the model receives a question $q_k$, the preceding conversation history $\mathcal{H}_{k-1} = \{(q_\ell, a_\ell)\}_{\ell=1}^{k-1}$, the clip observations $\mathcal{X} = \{I_{v,\tau}, K_{v,\tau}, E_{v,\tau}\}_{v,\tau}$, and the previous streaming slot $\mathbf{s}_{k-1}$. When geometric output is required, the model predicts a fixed-horizon world-coordinate trajectory $\hat{\mathcal{Y}}_k = \{\hat{\mathbf{p}}_{k,h}\}_{h=1}^{H}$ with $H=4$ and $\hat{\mathbf{p}}_{k,h} \in \mathbb{R}^{3}$.

\begin{figure*}[t]
\centering
\includegraphics[width=\textwidth]{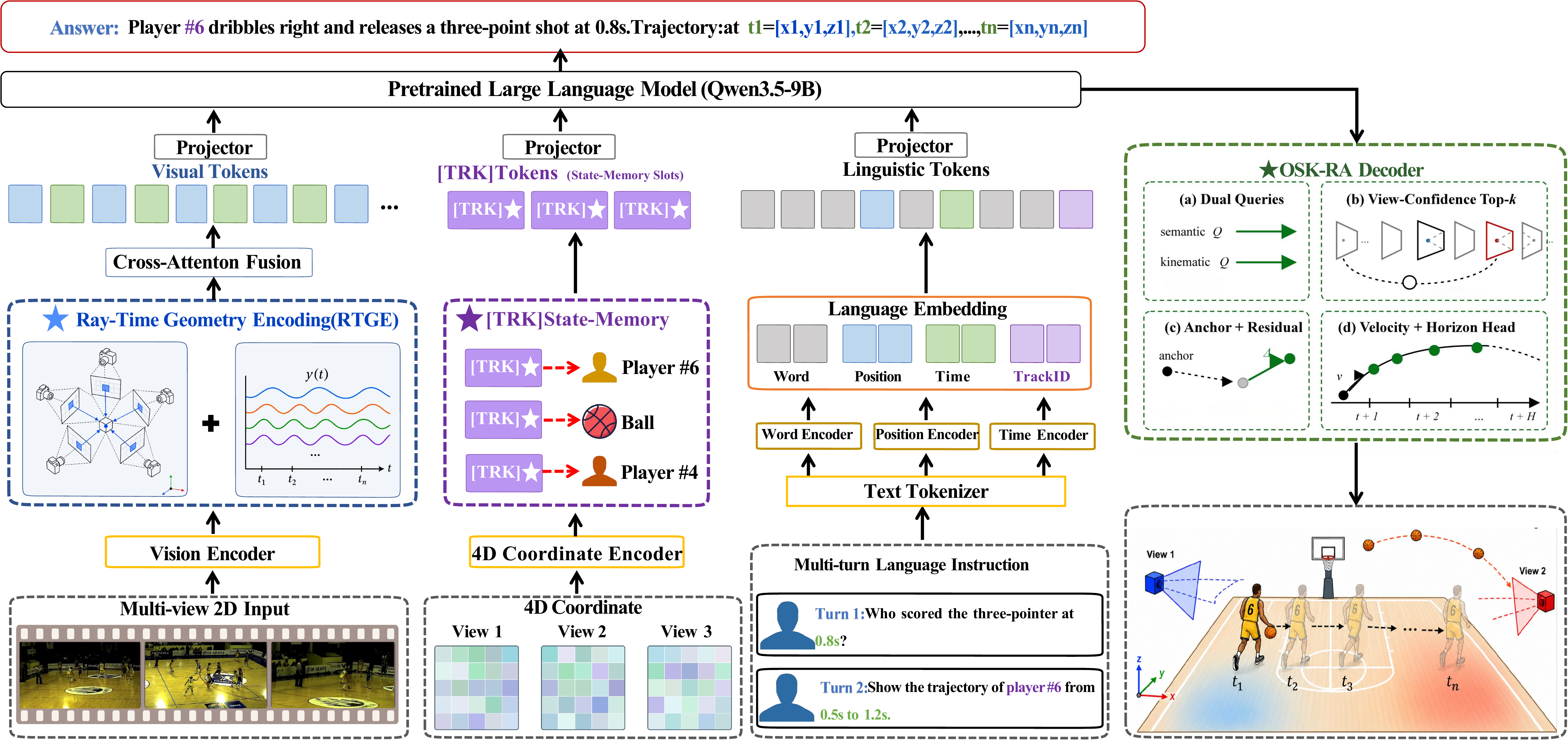}
\caption{Method overview of LMM-Track4D. RTGE injects ray-time geometry into visual tokens, TRK propagates target identity and motion evidence across dialogue turns, and OSK-RA decodes query-conditioned 3D trajectories from multiview observations. Coupled with language tokens from the multi-turn instruction, the system jointly answers dialogue queries and produces structured trajectories within a unified 4D dynamic reasoning pipeline.}
\label{fig:method_overview}
\end{figure*}

LMM-Track4D uses a Qwen3.5-9B backbone with Low-Rank Adaptation (LoRA) adapters~\cite{huLoRALowrankAdaptation2021} together with three trainable native modules: RTGE, a persistent TRK state-memory token, and the OSK-RA decoder. The remaining backbone weights remain frozen. The per-turn computation is
\[
\mathcal{F} = \Psi_{\mathrm{enc}}(\mathcal{X}), \qquad
(a_k, \mathbf{s}_k) = \Psi_{\mathrm{llm}}(q_k, \mathcal{H}_{k-1}, \mathcal{F}, \mathbf{s}_{k-1}), \qquad
\hat{\mathcal{Y}}_k = \Psi_{\mathrm{dec}}(\mathbf{s}_k, \mathcal{F}).
\]
Here $\mathcal{F}$ is the set of RTGE-conditioned visual tokens, $a_k$ is the language response, and $\mathbf{s}_k \in \mathbb{R}^{d}$ is the hidden state of the newly emitted TRK token. Text and trajectory outputs are produced in the same forward pass whenever geometric prediction is required. Figure~\ref{fig:method_overview} shows how these modules form an end-to-end pipeline for dialogue-grounded 4D dynamic reasoning.

\subsection{RTGE: Ray-Time Geometry Encoding}

RTGE is inserted between the visual encoder and the multimodal backbone. Let $\mathbf{f}_{v,\tau,n} \in \mathbb{R}^{d}$ denote patch $n$ extracted from image $I_{v,\tau}$ by the Vision Transformer (ViT). For the same patch, we compute the camera center $\mathbf{o}_{v,\tau} \in \mathbb{R}^{3}$ and the unit ray direction $\mathbf{d}_{v,\tau,n} \in \mathbb{R}^{3}$ from $K_{v,\tau}$ and $E_{v,\tau}$, and form the 6-D Pl\"ucker descriptor
\[
\mathbf{r}_{v,\tau,n} = [\mathbf{d}_{v,\tau,n}; \mathbf{m}_{v,\tau,n}] \in \mathbb{R}^{6}, \qquad
\mathbf{m}_{v,\tau,n} = \mathbf{o}_{v,\tau} \times \mathbf{d}_{v,\tau,n}.
\]
A learnable projection $\phi_r: \mathbb{R}^{6} \rightarrow \mathbb{R}^{d}$ maps the ray descriptor to the patch dimension. For time, we use a Fourier embedding
\[
\boldsymbol{\gamma}(\tau) = [\sin(\omega_i \tau), \cos(\omega_i \tau)]_{i=1}^{m},
\]
followed by a projection $\phi_t$ to the token dimension. The final RTGE-conditioned token is therefore
\[
\tilde{\mathbf{f}}_{v,\tau,n} = \mathbf{f}_{v,\tau,n} + \phi_r(\mathbf{r}_{v,\tau,n}) + \phi_t(\boldsymbol{\gamma}(\tau)),
\]
and the visual token set passed to the language backbone is
\[
\mathcal{F} = \{\tilde{\mathbf{f}}_{v,\tau,n}\}_{v,\tau,n}.
\]
Implementation-specific dimensional choices are deferred to Appendix~\ref{sec:appendix_impl_details}.

\subsection{TRK State-Memory Token}

TRK is a special token whose hidden state acts as the persistent target slot. At turn $k$, the backbone consumes the current query, dialogue history, RTGE-conditioned visual tokens, and previous slot state, and emits the answer together with a new slot state $\mathbf{s}_k$. This state is injected as the initial slot for turn $k+1$, so target identity and motion evidence accumulate across dialogue turns instead of being re-inferred from raw history each time. During training, cross-turn slot propagation uses stop-gradient, and detailed implementation settings are deferred to Appendix~\ref{sec:appendix_impl_details}.

\subsection{OSK-RA Decoder}

OSK-RA (Object-Slot Kinematic, Residual-Anchor) is the sole trajectory output head. It addresses the final 3D decoding bottleneck by separating target selection from geometric refinement. From the current TRK state, we form semantic and kinematic queries
\[
\mathbf{q}^{\mathrm{sem}}_k = W_{\mathrm{sem}}\mathbf{s}_k, \qquad
\mathbf{q}^{\mathrm{kin}}_k = W_{\mathrm{kin}}\mathbf{s}_k,
\]
where the semantic query identifies which target should be decoded and the kinematic query aligns the decoder with accumulated motion evidence. For each prediction step $h \in \{1, \ldots, H\}$, the decoder scores the retained evidence rays attached to RTGE-conditioned visual tokens,
\[
z_{k,h,r} = g^{h}_{\mathrm{score}}(\mathbf{q}^{\mathrm{sem}}_k, \mathbf{q}^{\mathrm{kin}}_k, \tilde{\mathbf{f}}_r), \qquad
\alpha_{k,h,r} = \operatorname{softmax}_r(z_{k,h,r}),
\]
where $r$ indexes the retained evidence rays $(\mathbf{o}_r, \mathbf{d}_r)$ associated with the selected tokens and $\sum_r \alpha_{k,h,r} = 1$. The fused step feature is
\[
\mathbf{z}_{k,h} = \sum_{r=1}^{R} \alpha_{k,h,r} \tilde{\mathbf{f}}_r.
\]
OSK-RA then computes a coarse multiview anchor by weighted line intersection,
\[
\hat{\mathbf{p}}^{\mathrm{anchor}}_{k,h}
=
\arg\min_{\mathbf{p} \in \mathbb{R}^{3}}
\sum_{r=1}^{R} \alpha_{k,h,r}
\left\|
(I - \mathbf{d}_r \mathbf{d}_r^{\top})(\mathbf{p} - \mathbf{o}_r)
\right\|_2^2,
\]
and refines it with a residual head,
\[
\Delta_{k,h} = g_{\mathrm{res}}([\mathbf{z}_{k,h}; \hat{\mathbf{p}}^{\mathrm{anchor}}_{k,h}]), \qquad
\hat{\mathbf{p}}_{k,h} = \hat{\mathbf{p}}^{\mathrm{anchor}}_{k,h} + \Delta_{k,h}, \qquad h = 1, \ldots, H.
\]
This anchor-plus-residual design reduces first-step bias accumulation, while the decoupled semantic and kinematic queries avoid forcing a single latent representation to carry both target identity and motion dynamics. Detailed view-selection and triangulation settings are deferred to Appendix~\ref{sec:appendix_impl_details}.

\subsection{Stage-wise Training and Objectives}

The optimization follows three stages while keeping the backbone frozen except for the LoRA adapters~\cite{huLoRALowrankAdaptation2021} and the native modules active in each stage. Stage 1 prioritizes trajectory supervision to stabilize geometry-grounded state estimation. Stage 2 jointly optimizes language generation and structured 3D prediction. Stage 3 applies group-relative policy optimization (GRPO)~\cite{shao2024deepseekmath} for lightweight policy alignment. Detailed schedules and systems settings are deferred to Appendix~\ref{sec:appendix_impl_details}.

During joint training, the objective combines assistant-token NLL with trajectory regression, temporal smoothness, and TRK identity-consistency supervision, so the model learns to answer spatiotemporal questions while maintaining coherent dynamic state across turns. Full loss formulas and training hyperparameters are given in Appendix~\ref{sec:appendix_impl_details}.

Overall, LMM-Track4D couples RTGE-conditioned geometric encoding, persistent TRK state propagation, and residual-anchor 3D decoding within a single multimodal dynamic reasoning framework.

\section{Experiments}
\label{sec:experiments}

The experiments address three questions: whether existing models can solve the task, whether Track4D-Bench diagnoses distinct 4D dynamic reasoning demands, and whether the gains of LMM-Track4D come from its dynamic-state design.

\subsection{Experimental Setup}

\paragraph{Protocol and metrics.} Experiments are reported under a unified dialogue-turn evaluation protocol on Track4D-Bench. The short regime evaluates full-clip trajectory recovery together with qualitative spatiotemporal dialogue, whereas the long regime evaluates local-state reasoning within a larger temporal window. All trainable models use the same scene-level train/validation/test partitions, with overlapping clips from the same source sequence kept within one split to avoid temporal leakage. At test time, each model receives synchronized keyframes and the current question, and all results are reported as single-run point estimates under the same prompting, parsing, and evaluation scripts. Reported metrics include CIDEr, BLEU-4, and METEOR~\cite{vedantam2015cider,papineni2002bleu,banerjee2005meteor} for non-trajectory dialogue turns, and SAcc@0.5, Traj-Acc@1.0, and TAcc@0.5 for explicit 3D outputs. All geometric metrics use meter-normalized coordinates, query-order alignment, and corpus-level micro averaging; detailed evaluation rules, decoding settings, and compute budgets are deferred to Appendix~\ref{sec:appendix_impl_details}.

\paragraph{Compared methods.} The comparison includes three groups of baselines: proprietary LMMs, open-source LMMs, and same-dataset SFT controls. The first two assess how far generic or 4D-oriented multimodal reasoning extends under a unified protocol, whereas the SFT controls span multiple backbone families and test whether stronger in-domain supervision alone can close the gap without explicit dynamic-state modeling. Table~\ref{tab:main_results} reports the full comparison.

\subsection{Comparison with Baselines}

The main comparison evaluates whether the task is already solved and whether Track4D-Bench reveals a substantive 4D dynamic reasoning gap.

\begin{table}[ht]
\caption{Complete baseline comparison on Track4D-Bench. Bold and underline denote the best and second best values, respectively; - indicates unavailable entries under the current protocol.}
\label{tab:main_results}
\centering
\small
\begin{tabular}{@{}lcccccc@{}}
\toprule
Method & C $\uparrow$ & B-4 $\uparrow$ & M $\uparrow$ & SAcc@0.5 $\uparrow$ & Traj-Acc@1.0 $\uparrow$ & TAcc@0.5 $\uparrow$ \\
\midrule
\multicolumn{7}{@{}l}{\textit{Proprietary LMMs}} \\
GPT-4o~\cite{hurst2024gpt} & 0.246 & 0.132 & - & - & - & - \\
Gemini 2.5 Pro~\cite{comanici2025gemini} & 0.155 & 0.062 & - & - & - & - \\
\midrule
\multicolumn{7}{@{}l}{\textit{Open-source LMMs}} \\
Qwen3-VL-9B-Instruct~\cite{bai2025qwen3} & 0.314 & 0.169 & - & - & - & - \\
LLaVA-Video~\cite{zhangLLaVAvideoVideoInstruction2024a} & 0.059 & 0.057 & 0.131 & - & - & - \\
Video-3D LLM~\cite{zhengVideo3DLLMLearninga} & 0.000 & 0.000 & 0.030 & - & - & - \\
LLaVA-3D~\cite{zhuLLaVA3DSimpleEffectivea} & 0.000 & 0.003 & 0.015 & - & - & - \\
MLLM-4D~\cite{yinMLLM4DVisualbasedSpatialtemporal2026a} & 0.133 & 0.094 & 0.164 & - & - & - \\
\midrule
\multicolumn{7}{@{}l}{\textit{Same-dataset SFT controls}} \\
Video-3D LLM + SFT~\cite{zhengVideo3DLLMLearninga} & 0.000 & 0.005 & 0.041 & - & - & - \\
LLaVA-Video + SFT~\cite{zhangLLaVAvideoVideoInstruction2024a} & 0.115 & 0.084 & 0.226 & - & - & - \\
Qwen2.5-VL-7B + SFT~\cite{Qwen2.5-VL} & 0.035 & 0.000 & 0.108 & - & - & - \\
Qwen3.5-9B + SFT~\cite{qwen3.5} & \underline{0.451} & \underline{0.220} & \underline{0.348} & - & - & - \\
\midrule
LMM-Track4D (Ours) & \textbf{0.698} & \textbf{0.322} & \textbf{0.453} & \textbf{0.758} & \textbf{0.626} & \textbf{0.619} \\
\bottomrule
\end{tabular}
\end{table}

Table~\ref{tab:main_results} indicates a clear gap. Strong proprietary and open-source LMMs achieve non-trivial language scores, but under a unified protocol they do not provide comparable structured 3D outputs, indicating that generic video reasoning does not solve trajectory-grounded dialogue. Broadening the same-dataset SFT controls likewise does not change the conclusion: although in-domain supervision improves several backbones, the strongest SFT control, Qwen3.5-9B + SFT, and the nominally 4D-oriented MLLM-4D both remain well below LMM-Track4D, which is the only model to combine strong language quality with robust geometry (0.758 SAcc@0.5, 0.626 Traj-Acc@1.0, 0.619 TAcc@0.5). The comparison therefore indicates an explicit dynamic-state advantage rather than a prompt-format or vanilla-SFT effect.

Table~\ref{tab:dataset_type_breakdown} separates performance by question type, clarifying where LMM-Track4D is already reliable and where continuous dynamic reasoning remains most demanding. Table~\ref{tab:history_dependence} then quantifies the extent to which this advantage depends on stable cross-turn state propagation.

\begin{table}[ht]
\caption{Question-type breakdown of LMM-Track4D on Track4D-Bench. Bold and underline denote the best and second best values, respectively. Entries marked with $^{\dagger}$ correspond to scene-level questions, for which language metrics are primary and object-centric spatial scores are not directly comparable.}
\label{tab:dataset_type_breakdown}
\centering
\small
\begin{tabular}{@{}lcccccc@{}}
\toprule
Type & C $\uparrow$ & B-4 $\uparrow$ & M $\uparrow$ & SAcc@0.5 $\uparrow$ & Traj-Acc@1.0 $\uparrow$ & TAcc@0.5 $\uparrow$ \\
\midrule
Identity & \textbf{1.274} & \underline{0.316} & \textbf{0.535} & 70.0 & 70.0 & 63.6 \\
Trajectory & \underline{0.678} & \textbf{0.335} & \underline{0.461} & \underline{87.9} & \underline{78.8} & \underline{77.8} \\
Relation & 0.539 & 0.221 & 0.446 & \textbf{100.0} & \textbf{81.1} & \textbf{89.9} \\
Motion trend & 0.420 & 0.154 & 0.378 & 78.8 & 51.5 & 44.4 \\
Scene-level & 0.369 & 0.202 & 0.416 & --$^{\dagger}$ & --$^{\dagger}$ & --$^{\dagger}$ \\
\bottomrule
\end{tabular}
\end{table}

This breakdown indicates that the benchmark is not reducible to a single-pattern task. Identity questions achieve the strongest language quality, while trajectory and relation questions produce the strongest geometry; relation peaks at 100.0 SAcc@0.5, 81.1 Traj-Acc@1.0, and 89.9 TAcc@0.5. Motion-trend remains the hardest category, indicating that updating state over time is still harder than recovering current identity, location, or relation. The split therefore provides a meaningful diagnostic of dynamic-reasoning subskills rather than a cosmetic question taxonomy.

\subsection{Ablation Study}

The ablation evaluates whether the gains arise from the intended dynamic-state modules rather than from backbone scale alone. Table~\ref{tab:ablation} compares the full model against two module removals that preserve the trajectory decoder, a TRK-free variant, and a backbone-only control. History effects are treated separately in Table~\ref{tab:history_dependence}.

\begin{table}[ht]
\caption{Component ablation of LMM-Track4D on Track4D-Bench. Bold and underline denote the best and second best values, respectively. Entries marked with $^{\dagger}$ do not yield stable structured geometric outputs under the current evaluation protocol, so only language quality is reported.}
\label{tab:ablation}
\centering
\small
\begin{tabular}{@{}lcccc@{}}
\toprule
Variant & C $\uparrow$ & SAcc@0.5 $\uparrow$ & Traj-Acc@1.0 $\uparrow$ & TAcc@0.5 $\uparrow$ \\
\midrule
Full model & \textbf{0.698} & \textbf{0.758} & \textbf{0.626} & \textbf{0.619} \\
w/o OSK-RA & \underline{0.631} & 0.588 & 0.500 & 0.425 \\
w/o RTGE & 0.579 & \underline{0.714} & \underline{0.571} & \underline{0.482} \\
w/o TRK & 0.509 & --$^{\dagger}$ & --$^{\dagger}$ & --$^{\dagger}$ \\
Backbone only & 0.451 & --$^{\dagger}$ & --$^{\dagger}$ & --$^{\dagger}$ \\
\bottomrule
\end{tabular}
\end{table}

The ablation results indicate a clear architectural contribution. Removing OSK-RA causes the largest drop in step-wise fidelity, reducing TAcc@0.5 from 0.619 to 0.425, while removing RTGE gives the second-largest degradation across the geometric metrics. Removing TRK, or collapsing to the backbone alone, eliminates stable structured geometry altogether. These patterns indicate that the gain is architectural rather than a simple backbone or training effect.

\subsection{History Dependence Analysis}

Table~\ref{tab:history_dependence} evaluates the extent to which continuous dynamic reasoning depends on carrying state across turns by comparing no-history, self-history, and gold-history settings across early, middle, and late turns.

\begin{table}[ht]
\caption{History dependence analysis under different turn-history sources.}
\label{tab:history_dependence}
\centering
\small
\begin{tabular}{@{}lcccccc@{}}
\toprule
& \multicolumn{3}{c}{Traj-Acc@1.0 $\uparrow$} & \multicolumn{3}{c}{TAcc@0.5 $\uparrow$} \\
\cmidrule(lr){2-4}\cmidrule(lr){5-7}
Source & Early & Middle & Late & Early & Middle & Late \\
\midrule
No history & 67.4 & 53.5 & 66.0 & 59.3 & 56.6 & 47.2 \\
Self history (ours) & 74.4 & 65.1 & 71.7 & 65.7 & 68.1 & 59.3 \\
Gold history (oracle) & 88.4 & 81.4 & 73.6 & 72.1 & 78.3 & 65.0 \\
\bottomrule
\end{tabular}
\end{table}

The history analysis indicates a clear continuity requirement. No-history performs worst, especially on late-turn TAcc@0.5, showing that current evidence alone is insufficient once the dialogue moves beyond the initial observation. Self history consistently improves over no history, while gold history remains best throughout. The remaining gap therefore comes mainly from imperfect cross-turn state propagation rather than from the decoder alone.

\subsection{Qualitative Comparison}

\begin{figure*}[t]
\centering
\includegraphics[width=\textwidth]{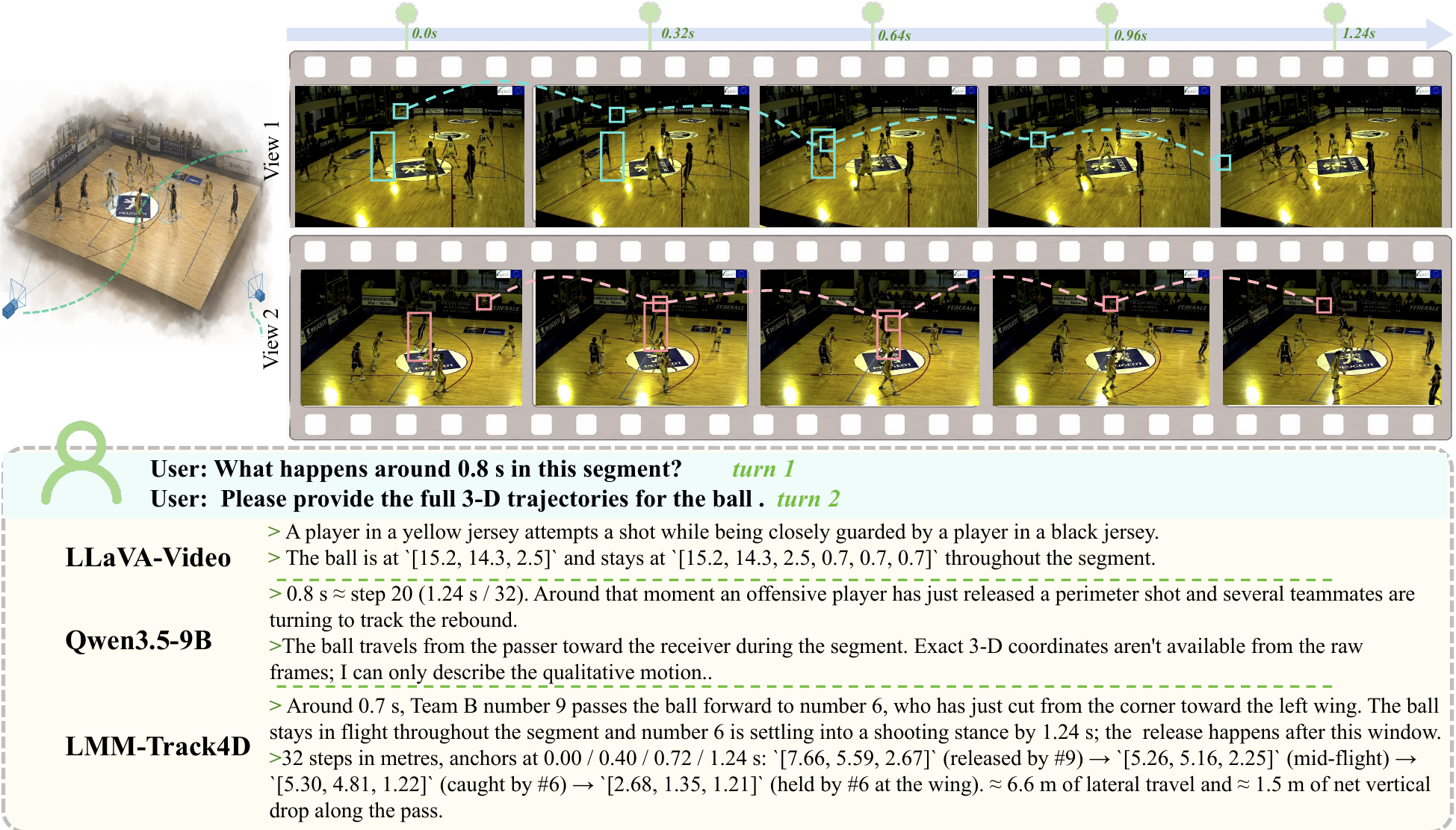}
\caption{Qualitative dialogue-grounded comparison. LMM-Track4D correctly identifies the shooter, localizes the release moment, and recovers the 3D trajectories of the player and ball. LLaVA-Video and Qwen3.5-9B are shown with dialogue responses only because they do not produce comparable structured trajectory outputs under the current protocol.}
\label{fig:qualitative}
\end{figure*}

Qualitative evidence is consistent with the quantitative trends. Figure~\ref{fig:qualitative} compares LMM-Track4D with LLaVA-Video and a Qwen3.5-9B baseline on a representative two-turn dialogue, illustrating the target capability: the model must resolve who acted, when the event occurred, and how the relevant entities evolve across time and views. Only LMM-Track4D is visualized with predicted trajectories, since the compared baselines do not produce stable structured trajectory outputs in this setting; other baselines are omitted for the same reason.

The main comparison shows that trajectory-grounded multi-turn spatiotemporal dialogue remains unsolved for current LMMs, while the diagnostic splits confirm that Track4D-Bench probes distinct dynamic-reasoning subskills. Ablation, history, and qualitative evidence then attribute the gains of LMM-Track4D to geometry grounding, persistent state propagation, and stable decoding, supporting explicit dynamic state modeling as an effective inductive bias for eliciting 4D dynamic reasoning.

\section{Conclusion}
\label{sec:conclusion}

We study a central limitation of current large multimodal models: they remain weak at reasoning over continuously evolving 4D dynamics across time, viewpoints, and multi-turn interaction. To address this gap, we formulate trajectory-grounded multi-turn spatiotemporal dialogue, introduce Track4D-Bench as a trainable and diagnostically interpretable benchmark, and present LMM-Track4D with RTGE, streaming TRK state propagation, and OSK-RA decoding. Results indicate that explicit dynamic state modeling is a practical route toward stronger 4D dynamic reasoning in LMMs, although the current setting still depends on structured supervision and remains limited in scale, temporal horizon, and openness of interaction.

\clearpage
\bibliographystyle{plainnat}
\bibliography{reference}

\end{document}